\documentclass[conference]{IEEEtran}
\IEEEoverridecommandlockouts

\usepackage{cite}
\usepackage{amsmath,amssymb,amsfonts}
\usepackage{algorithmic}
\usepackage{graphicx}
\usepackage{textcomp}
\usepackage{xcolor}
\usepackage{svg}
\usepackage{stfloats}
\usepackage{bbm}
\usepackage{multirow}
\def\BibTeX{{\rm B\kern-.05em{\sc i\kern-.025em b}\kern-.08em
    T\kern-.1667em\lower.7ex\hbox{E}\kern-.125emX}}
\begin{document}

\title{Improving Consistency in Vehicle Trajectory Prediction Through Preference Optimization
\thanks{\textsuperscript{1}Stellantis, Poissy, France; \textsuperscript{2}École des Mines de Paris, Paris, France.
}
}

\author{\IEEEauthorblockN{Caio Azevedo\textsuperscript{1, 2}, Lina Achaji\textsuperscript{1}, Stefano Sabatini\textsuperscript{1}, Nicola Poerio\textsuperscript{1}, \\ Grzegorz Bartyzel\textsuperscript{1}, Sascha Hornauer\textsuperscript{2}, Fabien Moutarde\textsuperscript{2}}
}


\maketitle

\begin{abstract}
Trajectory prediction is an essential step in the pipeline of an autonomous vehicle. Inaccurate or inconsistent predictions regarding the movement of agents in its surroundings lead to poorly planned maneuvers and potentially dangerous situations for the end-user. Current state-of-the-art deep-learning-based trajectory prediction models can achieve excellent accuracy on public datasets. However, when used in more complex, interactive scenarios, they often fail to capture important interdependencies between agents, leading to inconsistent predictions among agents in the traffic scene. Inspired by the efficacy of incorporating human preference into large language models, this work fine-tunes trajectory prediction models in multi-agent settings using preference optimization. By taking as input automatically calculated preference rankings among predicted futures in the fine-tuning process, our experiments--using state-of-the-art models on three separate datasets--show that we are able to significantly improve scene consistency while minimally sacrificing trajectory prediction accuracy and without adding any excess computational requirements at inference time.
\end{abstract}


\section{Introduction}

    Predicting the future behavior of agents in a given environment is essential for planning how to navigate that environment. In particular, for autonomous vehicles (AVs), it is a fundamental step to ensure safe driving, as the majority of crashes are caused by human error \cite{forbesstats}.

    The difficulty of the problem is its inherent stochastic nature, as we do not have access to the latent intentions of all agents. Before the advent of deep learning, physics-based methods were used to predict trajectories; these have since been largely surpassed by data-driven neural architectures which are able to implicitly learn a distribution over possible trajectories.

    Among data-driven predictors, there are those that do not explicitly model future interactions between different agents, called marginal predictors; and those that do, called joint predictors. In practice, marginal predictors generally achieve higher accuracy per agent than joint predictors since there is no need to also learn how to produce interdependent and internally consistent futures \cite{ngiamscene}; they also have simpler designs and consequently simpler training pipelines. Although they possess these advantages, simply applying them in interactive multi-agent scenarios can lead to inconsistent predictions among forecasted agents. These would include  collisions or inadequately scored predictions, i.e. when high likelihoods are assigned to predicted joint behaviors that deviate considerably from the ground-truth. Inconsistent predictions, when fed to the planning module in the typical AV pipeline, might lead to suboptimal decision-making in the form of unsafe driving due to the unexpected rollout of the scenario \cite{chen2022scept}.
    
    In the field of natural language processing, it has become standard practice \cite{achiam2023gpt, team2023gemini} to incorporate human preferences into the training pipeline of large language models (LLM) to ensure the safety of generations, a technique called preference optimization \cite{bai2022training, ouyang2022training}. In this work, we investigate if preference optimization can also be applied to trajectory prediction models to ensure \textit{scene consistency}, in a manner analogous to how it ensures safe text generations in LLMs.

    We identify some key parallels between LLMs and trajectory prediction models that motivate and justify the use of preference optimization for the latter class. For instance, both classes are inherently stochastic; however, unlike in LLMs, directly updating the predicted likelihoods to achieve desirable generations -- the goal of preference optimization -- has not been done in trajectory prediction models. Another parallel is that a human is generally capable of choosing which among two text generations is the better response, a feedback that is necessary for preference optimization algorithms; this is even more true for trajectory prediction scenarios, as a bad prediction is one that deviates too far from the ground truth or that has collisions, properties that can be detected automatically.

    In summary, this work addresses the question: \textit{ Can preference optimization be used to improve scene consistency in trajectory prediction models?}

    We try to answer this question by applying a recent preference optimization method, SimPO \cite{meng2024simpo} to improve the consistency of trajectory predictors when used in multi-agent scenarios. Our contributions are the following:

    \begin{enumerate}
        \item We propose a metric to automatically rank multiple prediction scenarios from a given input scene, a necessary step when applying preference optimization methods. 
        \item We show that an extended version of the SimPO objective can be applied to rankings of trajectory predictions output by any model, enhancing joint prediction consistency. We demonstrate this by presenting the positive effect of SimPO on collision rates in three different datasets, with only small accuracy degradations.
        \item We show that proposing multiple diverse candidate trajectories before applying SimPO can lead to further improvements in collision rates.
    \end{enumerate}

    Our work is the first to apply preference optimization to guide trajectory prediction models towards consistent predictions. It is a flexible fine-tuning strategy that can be applied to any preference metric, and any model that outputs future trajectories with associated likelihoods. Being a fine-tuning strategy, it is capable of considerably enhancing scene consistency without additional computational burdens at inference time and while keeping the overall shape of predicted trajectories realistic, as demonstrated by our experiments using state-of-the-art models in publicly available benchmarks.

    \begin{figure*}[t]
            \centering

        \includegraphics[width=\linewidth]{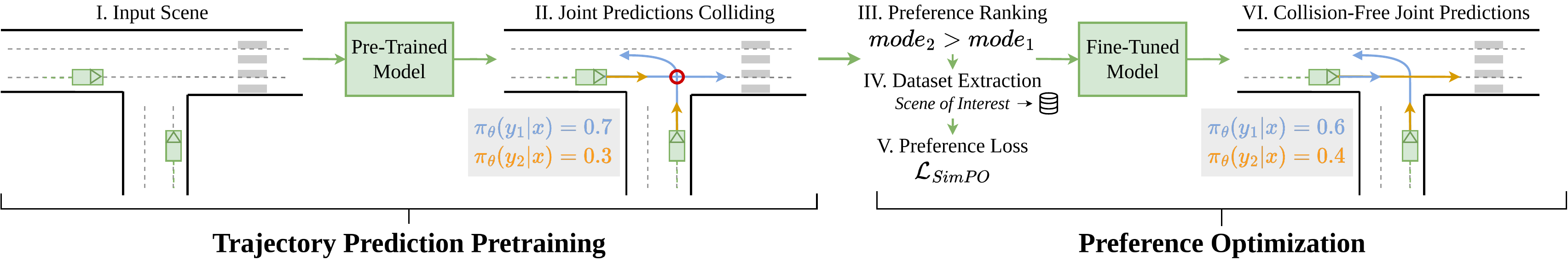}
        \caption{Illustration of our method on a hypothetical scenario. The input scene (I) is processed by a model that outputs unrealistic interactions, such as collisions (II). We aggregate these into joint predictions and rank the different modes according to a preference metric (III). In this work, mode 2 is preferred over 1 because it does not contain collisions. We extract a preference dataset containing interesting scenes for later fine-tuning (IV). The ranking and joint likelihoods make up the loss function during preference optimization (V). The result after fine-tuning are correctly paired predictions with updated likelihoods (VI).}
        \label{fig:simpo_schematics}
    \end{figure*}

\section{Related Work}
\label{sec:related_work}

    \paragraph{Marginal predictors}
    Marginal trajectory predictors decode future trajectories independently for each agent in the scene, without explicitly modeling future interactions between agents. They can be broadly categorized based on how future trajectories are decoded. LaneGCN \cite{liang2020learning} directly decodes from map-aware agent encodings multi-modal predictions through an MLP. More recent works utilize an anchor-based decoding framework, in which either fixed \cite{shi2022motion, chai2020multipath} or adaptive \cite{varadarajan2022multipathpp, nayakanti2023wayformer, zhou2023query, lin2024eda} anchors define mixture model components able to capture both intent and control uncertainty \cite{chai2020multipath}. Among these, QCNet \cite{zhou2023query} employs a DETR-like decoder \cite{carion2020end} that lets a set of learnable mode queries attend to scene context to output adaptive anchors which are then refined into predicted trajectories. In our work we perform experiments with both classes of models by using the baseline version of FJMP \cite{rowe2023fjmp} (which builds on LaneGCN) and QCNet, showing how both classes of marginal models can benefit from preference optimization and effectively output consistent joint predictions after the fine-tuning process. 

    \paragraph{Joint predictors} Joint trajectory prediction methods \cite{ngiamscene, sun2022m2i, rowe2023fjmp, zhou2023qcnext, shi2024mtr++, liu2024reasoning} adopt various strategies to ensure scene consistency. Interaction-graph-conditioned approaches \cite{sun2022m2i, rowe2023fjmp, liu2024reasoning} rely on predicted interaction edges between agents to restrict what each agent attends to \cite{liu2024reasoning} or to sequentially decode conditional trajectories \cite{sun2022m2i, rowe2023fjmp}. FJMP \cite{rowe2023fjmp}, in this latter class, achieved state-of-the-art performance on the Interaction dataset \cite{interactiondataset}. It introduces a module that predicts a directed acyclic graph (DAG) of agent interactions, enabling conditional trajectory generation based on the DAG's partial ordering.
    Though this effectively models interactive scenarios, this process adds considerable computational overhead due to multi-stage training and sequential decoding.
    BeTop \cite{liu2024reasoning} uses interaction topology-guided trajectory decoding, using braid theory to construct interaction labels. It showed considerable improvement on the Waymo Open Motion Dataset's Interaction Challenge \cite{ettinger2021large}.
    %
    Our work is capable of fine-tuning any marginal predictor into outputting consistent joint predictions using preference optimization.
    This approach retains the advantages of marginal models and dispenses with the computational requirements of joint models such as FJMP. We also show that we can apply SimPO to a joint model such as BeTop and get further improvements in scene consistency.

    \paragraph{Preference Optimization} After the successful application of reinforcement learning with human feedback (RLHF) \cite{ziegler2019fine} to LLMs \cite{bai2022training, ouyang2022training}, many approaches have been proposed \cite{yuan2024rrhf, ethayarajh2024kto, azar2024general, zhao2023slic, rafailov2024direct} to simplify its multiple steps or avoid high memory costs, without degrading final generation quality. DPO \cite{rafailov2024direct} demonstrates that the objective function of RLHF can be optimized directly, without the need to train a separate reward model. SimPO \cite{meng2024simpo}, the method our work is based upon, further simplifies the optimization algorithm by dispensing the need of a supervised fine-tuned model to serve as a reference in the objective function, and aligning the optimized rewards with the metric used during inference. SimPO achieves impressive results when compared to many other preference optimization objectives with an elegantly simple algorithm, inspiring us to adapt it for our use case. As far as we know, we are the first to apply preference optimization to trajectory prediction.

\section{Preliminaries}
\label{sec:preliminaries}

    \subsection{Trajectory Prediction}

        A trajectory prediction scene is the combination of a high-definition map (HD-Map), usually represented as a lane graph, and agent features such as  position $\mathbf{p}_i^t = (p_{i, x}^t, p_{i, y}^t)$, velocity $\mathbf{v}_i^t = (v_{i, x}^t, v_{i, y}^t)$ and yaw angle $\psi_i^t$, for agent $i \in \{ 1, ..., A \}$ and across past timesteps $t \in \{ -T_{obs}+1, ..., 0 \}$. The goal is to predict the future positions of a single or multiple agents in the scene for a certain future time horizon $\{ 1, ..., T_{\text{fut}} \}$.

        Trajectory prediction benchmarks therefore usually require models to generate several predictions for the same agent so that ideally the different possible maneuvers the agent could adopt are taken into account in the planning phase. In a multi-agent setting, we say that predictions happening simultaneously amongst all agents belong to the same traffic scene mode.

    \subsection{Preference Optimization}

        \paragraph{Problem Formulation} Preference optimization comes originally from cognitive neuroscience and relies on the idea that for example, given a choice between two objects, the probability that a human would prefer one of them instead of the other depends on an implicit reward function~$r^*$. More concretely, in the case of LLMs, the probability $p(y_1 \succ y_2 | y_1, y_2, x)$ that a human would prefer generation $y_1$ instead of $y_2$ given an input prompt $x$ depends on the values of $r^*(x, y_1)$ and $r^*(x, y_2)$. Of course, what will decide whether $y_1$ is preferred compared to $y_2$ depends on the specific goal being pursued.

        The aim is to enable models to generate outputs with higher rewards, which  translates to higher-quality outputs according to the target task.

        \paragraph{Preference Distributions} Two models for the preference distribution based on the implicit reward function have been proposed: the Bradley-Terry (BT) \cite{bradley1952rank} model, which deals with comparisons between two outputs; and the Plackett-Luce (PL) \cite{plackett1975analysis, luce1959individual} models, which extends the BT model to rankings of outputs.

        The BT model assumes the following expression for the preference distribution:
        \begin{equation}
            \small
            p^*(y_1 \succ y_2 | y_1, y_2, x) = \frac{\exp (r^*(x, y_1))}{\exp (r^*(x, y_1)) + \exp (r^*(x, y_2))}.
            \label{eq:bt_model}
        \end{equation}
        Thus the higher the implicit reward of $y_1$ the more likely it is to be preferred over $y_2$.
        
        In the PL model, we consider a function $\tau: \{1, ..., K\} \rightarrow \{1, ..., K\}$ that takes as input the relative rank of an output and maps it to its corresponding index (thus $\tau(1)$ is the index of the best output). The preference distribution is then \cite{rafailov2024direct}
        \begin{equation}
            \small
            p^*(\tau | y_1, ..., y_K, x) = \prod_{k=1}^{K} \frac{\exp (r^*(x, y_{\tau (k)}))}{ \sum_{j=k}^K \exp (r^*(x, y_{\tau (j)})) }.
            \label{eq:pl_model}
        \end{equation}
        Note that when $K=2$ the expression is reduced to (\ref{eq:bt_model}). The intuition is that output $\tau(k)$ has its reward compared to that of outputs $ \tau(k+1), ..., \tau(K)$, i.e. the outputs lower in ranking. If these rewards are much lower than that of $\tau(k)$ for most $k$ it means that the ranking given has high likelihood: a probable ranking will most likely have its rewards ordered accordingly.
        
        \paragraph{Optimization} Given (\ref{eq:bt_model}) and (\ref{eq:pl_model}), one can now parameterize the reward function by $r_\theta$ and minimize the negative log-likelihood. The objective function for the BT model is then
        \begin{equation}
            \small
             \mathcal{L} (\theta) = - \mathbb{E}\left[ \log \sigma( r_\theta (x, y_w) - r_\theta (x, y_l) )  \right],
             \label{eq:bt_loss1}
        \end{equation}
        with $y_w$ and $y_l$ being the preferred and dispreferred outputs respectively. For the PL model, the loss is given by
        \begin{equation}
            \small
             \mathcal{L} (\theta) = - \mathbb{E}\left[ \log \prod_{k=1}^K \frac{\exp (r_\theta(x, y_{\tau (k)}))}{ \sum_{j=k}^K \exp (r_\theta(x, y_{\tau (j)})) } \right]. 
             \label{eq:pl_loss1}
        \end{equation}
        Here, the expectation is taken over ${\tau, y_1, ..., y_K, x \sim \mathcal{D}}$. We explain in more detail our reward formulation in the specific setting of trajectory prediction in Section \ref{sec:simpo_objective}.

        \paragraph{Simple Preference Optimization}
        SimPO \cite{meng2024simpo} models the reward of a generation $y$ by its predicted log-likelihood $\pi_\theta (y|x)$:
        \begin{equation}
            \small
            r_\theta (x, y) = \frac{\beta}{|y|} \log \pi_\theta (y | x).
            \label{eq:simpo_llm_reward}
        \end{equation}
        With $|y|$ being the token-length of text generations for LLMs. This way, the reward being optimized directly corresponds to the metric used in the generation process, leading to a better correlation between high-reward outputs and desired responses. The normalizing factor $|y|$ is empirically shown to help avoid longer but lower-quality generations, and $\beta$ is a hyperparameter. Finally, a target reward margin $\gamma$ is added in the loss from (\ref{eq:bt_loss1}), so that
        \begin{equation}
            \small 
            \begin{split}
                \mathcal{L}_{\text{SimPO}}(\theta) = &  -\mathbb{E} \left[ \log  \sigma \left( \frac{\beta}{|y_w|} \log \pi_\theta (y_w | x) \right. \right. \\
                & \left. \left. - \frac{\beta}{|y_l|} \log \pi_\theta (y_l | x) - \gamma \right) \right],
            \end{split}
            \label{eq:simpo_llm_loss}
        \end{equation}
        with $y_w, y_l, x \sim \mathcal{D}$. The purpose of $\gamma$ is to force rewards of winning responses to be at least $\gamma > 0$ higher than rewards of losing responses. Its use is empirically shown to lead to much better outputs. SimPO presents a truly simple pipeline to perform preference optimization without the need of extra supervised fine-tuning steps or separate reference models, while achieving the same level of improvement as other methods.

\section{Methodology}
\label{sec:method}

    Fig. \ref{fig:simpo_schematics} showcases our contribution. In a first step, we apply (any) pretrained trajectory prediction model on an input scene with past agent trajectories and map information. The resulting agent-level predicted trajectories are likely to contain collisions after likelihood-based aggregation into scene-level joint predictions.
    
    The second step is the core of this work: in Section \ref{sec:pref_metric} we show how we are able to automatically establish a ranking of the multi-agent modes by proposing a collision-penalizing preference cost. In Section \ref{sec:pref_data_extraction} we suggest a way to extract a subset of the training set containing interactive scenes, to lessen the computational requirements of the fine-tuning step and limit noise in its samples. Finally, in Section \ref{sec:simpo_objective} we propose an adaptation of the SimPO \cite{meng2024simpo} objective for trajectory prediction to be used in fine-tuning. By updating the  model in this way, inconsistent behaviors in the new joint predictions are effectively prevented.

    \subsection{Preference Metric}
    \label{sec:pref_metric}

        Judging the quality of a joint prediction at training time is a straight-forward task: modes with collisions between agents are necessarily worse, as are ones that jointly deviate too much from the ground-truth. This fact allows us to automate a necessary step in applying preference optimization for LLMs: labeling preferred generations in pairwise comparisons. 
        
        We do this by using two elements: the average FDE of each mode, added with a collision detection metric. In particular, for detecting collisions, we use for mode $k$ the repeller cost $R_k$ from MotionDiffuser \cite{jiang2023motiondiffuser}:
        \begin{equation}
            \small
            \mathbf{A}_k = \max \left\{ \left( 1 - \frac{\mathbf{\Delta}_k}{r} \right) \odot (1 - \mathbf{I}), \mathbf{0} \right\},
            \label{eq:dist_matrix}
        \end{equation}
        \begin{equation}
            \small
            R_k = \frac{\sum \mathbf{A}_k}{\sum (\mathbf{A}_k > 0) + \epsilon}.
            \label{eq:repeller_cost}
        \end{equation}
        Here, $\mathbf{\Delta}_k \in \mathbb{R}^{A \times A \times T_{\text{fut}}}$ is the pairwise L2 distance between agents at each timestep of mode $k$, $\mathbf{I}$ is the identity tensor broadcast to all timesteps, and $r$ is a hyperparameter that controls the minimum distance between agents to count an interaction. The $\epsilon$ term simply avoids division by zero.

        The total preference cost $C_k$ for each mode is the sum of its average final displacement error with its weighted repeller cost:
        \begin{equation}
            \small
            C_k = \text{avgFDE}_k + \lambda R_k.
            \label{eq:preference_metric}
        \end{equation}
        Setting a high value for $\lambda$ guarantees that modes with collisions will be pushed down in the preference ranking. The lower the value of the cost $C_k$, the better mode $k$ is. It is important to note, however, that any method that outputs a ranking between different future predicted modes can in principle be applied.

    \subsection{Preference Dataset Extraction}
    \label{sec:pref_data_extraction}

        The dataset we use for preference optimization, $\mathcal{D}$, is a  subset of the training set of the benchmarks we experiment on. This subset consists of either scenes with collisions, i.e. agents come closer than $1m$ to each other, or scenes satisfying $C_{max} - C_{min} > \delta$, with $C_{max} = \max_i \{ C_i \}_{i=1}^K$ and analogously for $C_{min}$, $\delta$ being a hyperparameter. The intuition behind the latter criterion is that if all modes had too similar a value of preference cost, such that the ranking would become arbitrary, noise would possibly destabilize the training process. The difference between the preference dataset of trajectory prediction and LLMs is that the ranking between modes for each sample is automatic, not requiring expensive manual labelling; and also that generations themselves are not in the dataset but are allowed to be updated dynamically, since trajectory prediction models generally cannot, due to their design, calculate the likelihood of a fixed in-dataset future prediction.

    \subsection{Preference Optimization for Multi-Agent  Prediction}
    \label{sec:simpo_objective}

        The goal of applying preference optimization for trajectory prediction is that, by updating the scores of joint predictions, modes containing collisions will be pushed down in the preference ranking, which is equivalent to having their likelihood decreased. We expect, therefore, decreases in the preference loss to lead to either lower assigned probabilities for modes with collisions or different joint mode aggregations that avoid collisions. 
        
        To achieve this, we adapt the SimPO objective in (\ref{eq:simpo_llm_loss}) to trajectory prediction, by first noting that marginal models normally decode agent-level likelihoods, so that it becomes necessary to aggregate them into scene-level likelihoods. We do this by pairing the agents' predictions by order of likelihood: the most likely prediction of all agents make up the first scene mode, and so on. This straightforward aggregation strategy is not a requirement of our method, but simply ensures a natural and interpretable composition of agent-level predictions into joint ones while minimizing computational overhead. By adopting this aggregation strategy our method scales to complex scenes with many agents, even when using a high value of $K$ predicted trajectories per agent; it allows us to focus on the effects of preference optimization. Once the pairings are done we can simply average the logits of all agents from the same pairing to get $K$ scene-level likelihoods $\pi_\theta(y_k | x)$.
        
        Furthermore, in our case generations are always the same length $T_\text{fut}$, and likelihoods are given to an entire mode directly. For these reasons we may remove the length-normalizing factor $|y|$:
        \begin{equation}
            \small
            r_\theta (x, y) = \beta \log \pi_\theta (y | x).
            \label{eq:simpo_tp_reward}
        \end{equation}
        Next, to leverage the full ranking of $K$ modes in the loss, we adapt it to use the PL model instead. By substituting (\ref{eq:simpo_tp_reward}) into (\ref{eq:pl_loss1}), we get
        \begin{equation}
            \small
            \mathcal{L}_{\text{SimPO}} (\theta) = - \mathbb{E}\left[ \log \prod_{k=1}^K \frac{ \exp({ \beta \log \pi_\theta (y_{\tau (k)} | x )  } ) }{ \sum_{j=k}^K  \exp({  \beta \log \pi_\theta (y_{\tau (j)} | x )  } )} \right].
            \label{eq:simpo_tp_loss_no_gamma}
        \end{equation}
        Finally, to adapt the target reward margin $\gamma$, we add a term that scales with the difference in relative position of each mode in the preference ranking:
        \begin{equation}
            \small
            \mathcal{L}_{\text{SimPO}} (\theta) = - \mathbb{E}\left[ \log \prod_{k=1}^K \frac{ \exp({ \beta \log \pi_\theta (y_{\tau (k)} | x ) + k \gamma }) }{ \sum_{j=k}^K  \exp({  \beta \log \pi_\theta (y_{\tau (j)} | x ) + j \gamma }) } \right] .
            \label{eq:simpo_tp_loss}
        \end{equation}


\section{Experiments}
\label{sec:exp}

    \subsection{Experimental Setup}
    \label{sec:exp_metrics}

        \paragraph{Datasets.} Many datasets are available to compare different trajectory prediction models. In particular, for our experiments, we utilize Argoverse 2 (AV2) \cite{wilson2021argoverse}, Interaction \cite{interactiondataset}, and the joint prediction split of the Waymo Open Motion Dataset (WOMD) \cite{ettinger2021large}. These come with their own HD-maps of different locations, and many scenes with fully-annotated agent past trajectories and properties such as type or size.

        \paragraph{Metrics.}
        Under a multi-agent prediction setting, we consider a certain mode to be \textit{consistent} in a strict sense if its predicted future trajectories do not collide with each other. Let $c_k$ be the number of collisions detected in mode $k$. The scene collision rate (SCR) is the ratio of modes that contain collisions in a scene, aggregated across the dataset:
        \begin{equation}
            \small
            SCR=\frac{1}{K} \sum_{k=1}^K \mathbf{1}_{c_k > 0}.
            \label{eq:scr_def}
        \end{equation}
        This metric is similar to ones used in public benchmarks such as Interaction \cite{interactiondataset}. However, even if a mode has collisions, it could be the case that the predicted likelihood of that mode is close to zero. In this case the set of predicted modes can still be called, in a looser sense, consistent. To measure this we introduce a new metric, probability-weighted scene collision rate (pSCR), that corresponds to the expected ratio of modes with collisions under the predicted trajectory distribution. For a single scene, it is given by
        \begin{equation}
            \small
            pSCR= \sum_{k=1}^K \pi_\theta (y_k | x) \mathbf{1}_{c_k > 0}.
            \label{eq:pscr_def}
        \end{equation}
        As is done in AV2, we consider that a collision is detected if two agents come closer than 1m to each other in the same mode $k$ at any timestep.

        We expect that pSCR be more correlated to the SimPO loss in (\ref{eq:simpo_tp_loss}), as modes with collisions will ideally have their likelihoods decreased since they are low in the preference ranking.
        
        It is essential that the model retains accuracy after applying preference optimization by predicting close to the ground-truth. We measure this with MinJointFDE$_K$, which is calculated by taking the FDE (distance between predicted and ground-truth future end point) of all the agents in a single mode, averaging them and then taking the minimal value among the modes. MinJointFDE$_K$ on WOMD is calculated only at the final 8s timestep.
        All metrics, including collision rates, are calculated for most probable six joint modes.

        \paragraph{Models.} For our experiments, we utilize four models which are publicly available and state-of-the-art at their respective benchmarks at the time of release: QCNet \cite{zhou2023query}, a marginal predictor available for AV2; modified versions (to add scoring of trajectories--this change explains the small discrepancies between initial values in Tables \ref{tab:main_results_marginal} and \ref{tab:main_results_joint} and reported metrics in the original paper) of both the baseline non-factorized version of FJMP \cite{rowe2023fjmp}, i.e. not using the predicted future interaction graphs (which we call FJMP-Marginal), and its full version (FJMP-Joint), available for both AV2 and Interaction; and BeTop \cite{liu2024reasoning}, available for WOMD.

        \paragraph{Implementation Details.} We apply 5 epochs of SimPO on the preference dataset $\mathcal{D}$, which contains around 15\% of the training set, and was extracted using $\delta = 2.5$ for AV2 and $\delta = 1.0$ for Interaction, since the latter is an easier benchmark. For the results in Tables \ref{tab:main_results_marginal} and \ref{tab:main_results_joint}, we use $\beta = 2$ in all cases, $\gamma = 5$ for QCNet and BeTop and $\gamma = 15$ for FJMP-Marginal, and $\lambda = 10^3$ for the preference metric, ensuring that collision modes are always pushed down in ranking. The learning rate used was $10^{-5}$ for all runs. In Section \ref{sec:ablation} we present the impact of each hyperparameter on final results.

    \begin{table}[t]
        \caption{Main results of applying SimPO on marginal models,  $K = 6$. Collision rates are multiplied by $10^{3}$.}
        \centering
        \begin{tabular}{@{}l@{\hspace{2pt}}|l|l@{\hspace{1pt}}l@{\hspace{1pt}}l@{}}
        \hline
        & Model               & SCR          & pSCR            & MinJointFDE$_6$ \\ \hline \hline

        \multirow{2}{*}{\rotatebox{90}{\scriptsize  Int}} & FJMP-Marginal              &  0.16   & 0.10        &  0.616           \\
        & FJMP-Marginal + SimPO & 0.14 \textcolor{blue}{\scriptsize (-9\%)}  & 0.09 \textcolor{blue}{\scriptsize (-7\%)}  & 0.617 \textcolor{red}{\scriptsize (+0.2\%)}    \\
        \hline
       
        \multirow{4}{*}{\rotatebox{90}{\scriptsize  AV2}} & {FJMP-Marginal} & {28.43}  & {25.43}   & {2.128}   \\
        & {FJMP-Marginal+ SimPO} & {26.52 \textcolor{blue}{\scriptsize (-7\%)}} & {14.29\textcolor{blue}{\scriptsize (-44\%)}} & {2.199 \textcolor{red}{\scriptsize (+9\%)}}  \\ 
        \cline{2-5}
        &{QCNet}               & {8.64}         & {2.86}         & {1.366}            \\
        & {QCNet + SimPO} & {6.29 \textcolor{blue}{\scriptsize (-27\%)}} & {1.23 \textcolor{blue}{\scriptsize (-57\%)}} & {1.375 \textcolor{red}{\scriptsize (+1\%)}}   \\ 
        
        \hline
        
        \end{tabular}
        
        \label{tab:main_results_marginal}
    \end{table}

    \begin{table}[t]
        \caption{Main results of applying SimPO on joint models,  $K = 6$. Collision rates are multiplied by $10^{3}$.}
        \centering
        \begin{tabular}{@{}l@{\hspace{2pt}}|l|l@{\hspace{1pt}}l@{\hspace{1pt}}l@{}}
        \hline
        & Model               & SCR          & pSCR            & MinJointFDE$_6$ \\ \hline \hline

        \multirow{2}{*}{\rotatebox{90}{\scriptsize  Int}} & FJMP-Joint    &  0.10  & 0.10   &  0.633               \\
        & FJMP-Joint + SimPO & 0.08 \textcolor{blue}{\scriptsize (-14\%)} & 0.09 \textcolor{blue}{\scriptsize (-3\%)}  &  0.745 \textcolor{red}{\scriptsize (+17\%)}          \\
        \hline
       
        \multirow{2}{*}{\rotatebox{90}{\scriptsize AV2}} & {FJMP-Joint}  & {25.34}  & {29.40}  & {2.013}     \\
        & {FJMP-Joint + SimPO}  & 25.20   \textcolor{blue}{\scriptsize (-0.5\%)}  & 27.83 \textcolor{blue}{\scriptsize (-5\%)}  & 2.020 \textcolor{red}{\scriptsize (+0.3\%)}        \\

        \hline

        \multirow{2}{*}{\rotatebox{90}{\scriptsize Way}} & BeTop             & 12.81   & 11.55  &  4.153   \\
        & BeTop + SimPO & 8.14 \textcolor{blue}{\scriptsize (-36\%)}  & 7.24 \textcolor{blue}{\scriptsize (-37\%)}  & 4.453 \textcolor{red}{\scriptsize (+7\%)}    \\ 
        
        \hline
        
        \end{tabular}
        
        \label{tab:main_results_joint}
    \end{table}

        \begin{figure}
            \centering
            

            \includegraphics[width=0.9\linewidth]{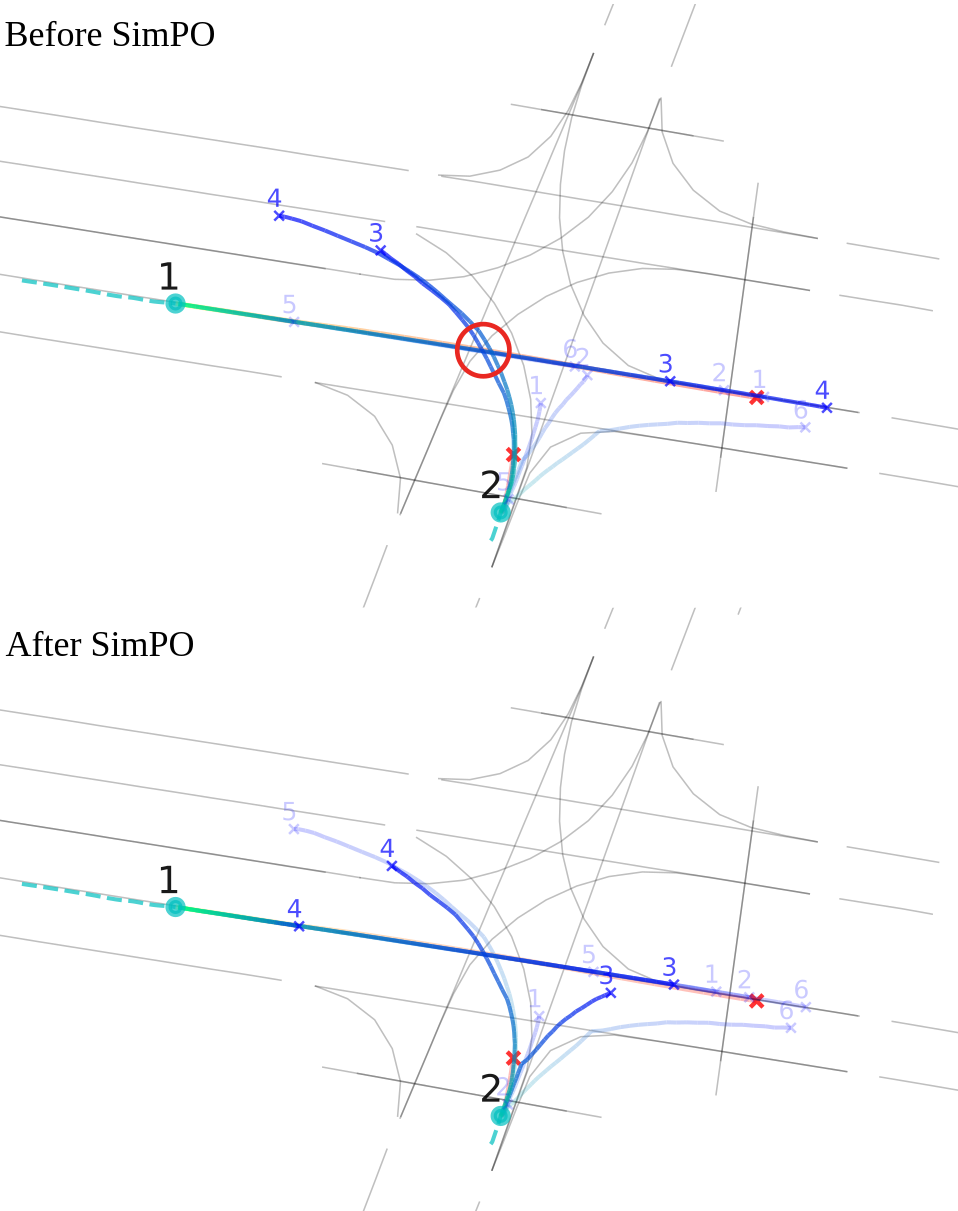}
            \caption{Qualitative example of SimPO effects, $K=6$. Ground-truth future is shown in red and past trajectories in dashed light blue. Predictions with the same number belong to the same mode. Modes 3 and 4 are darker to highlight effect of SimPO. Before, there is a collision between agents 1 and 2 at modes 3 and 4 within the red circle. After, no collisions take place as aggregation changed. Unrealistic modes have decreased probability, e.g. modes 5 and 6. }
            \label{fig:k6_qcnet_qualitative_case}
        \end{figure}

        \begin{figure}[t]
            \centering
            \includegraphics[width=0.95\linewidth]{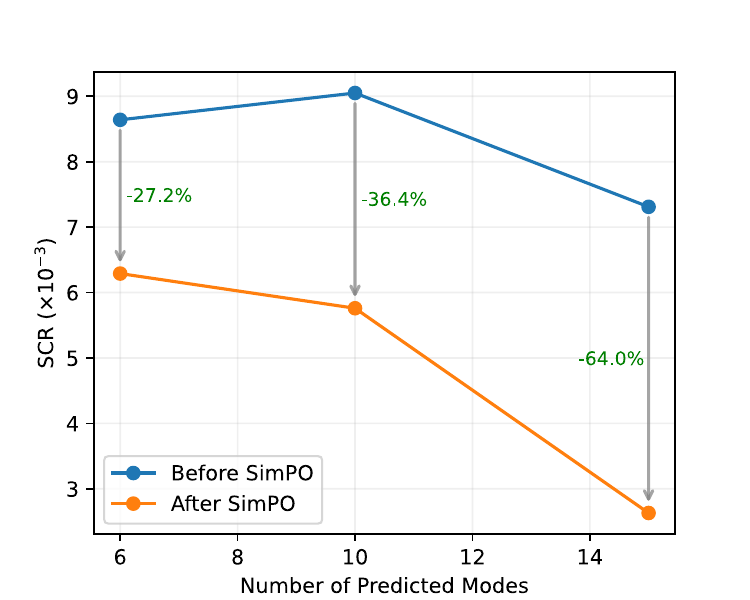}
            \caption{Effectiveness on QCNet of oversampling, applying SimPO and selecting top 6 modes.}
            \label{fig:oversampling_scr}
        \end{figure}

        \begin{figure}[t]
            \centering
            
            

            \includegraphics[width=\linewidth]{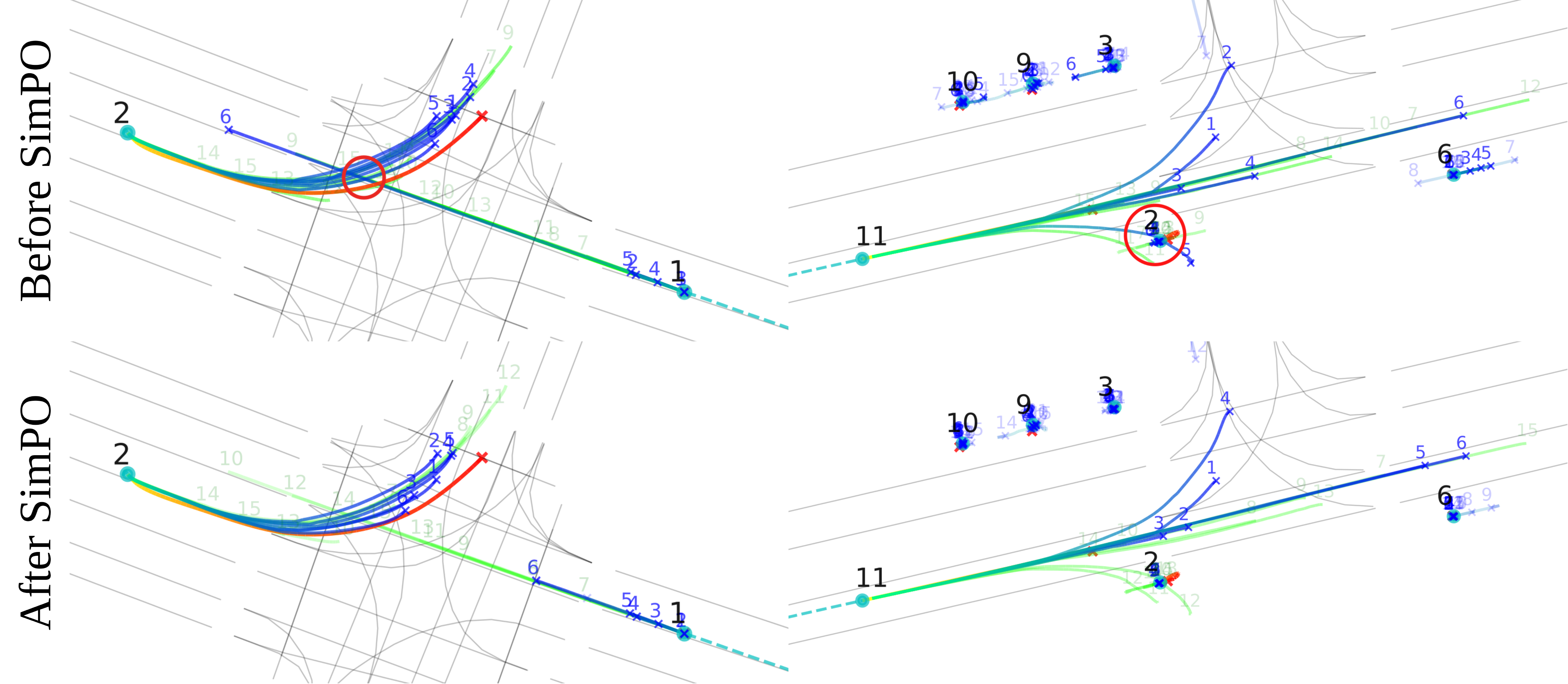}
            
            \caption{Qualitative example of the effectiveness of SimPO combined with oversampling, using QCNet and $K=15$.  Most likely six modes, which are used in evaluation, are in dark blue; others in light green. Collisions occur inside red circles. After~SimPO, notice that inconsistent interactions are mitigated, and trajectory shapes are kept realistic.}
            \label{fig:k15_qcnet_qualitative_case}
        \end{figure}

        \begin{figure}[t]
            \centering
            \includegraphics[width=0.7\linewidth]{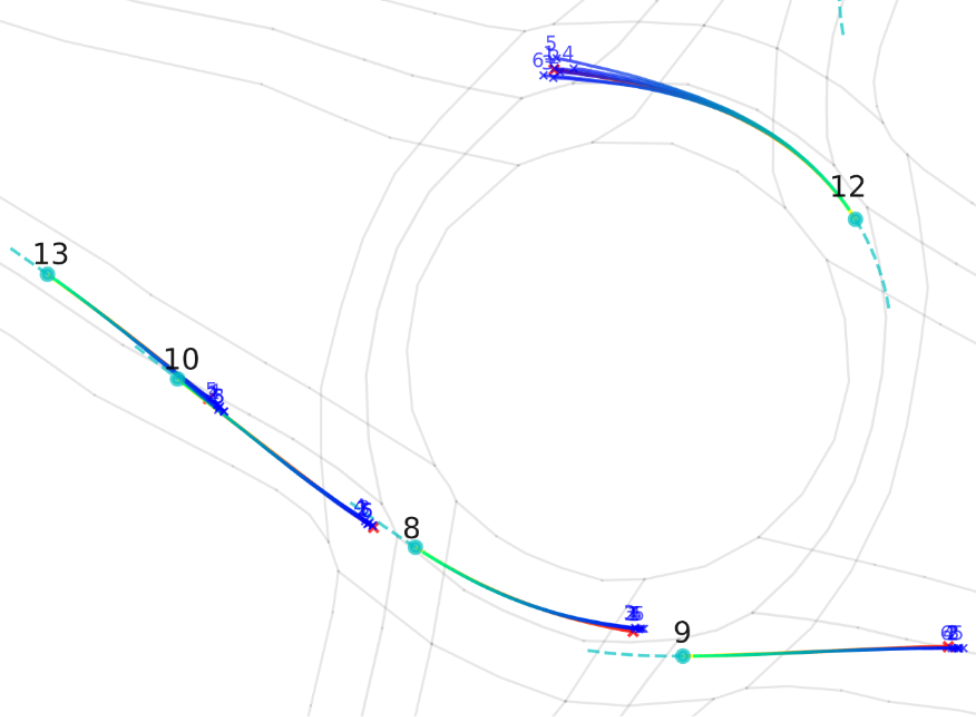}
            \caption{Example of FJMP-Marginal predictions to demonstrate mode collapse in the Interaction dataset. FJMP-Joint suffers from similar mode collapse.}
            \label{fig:fjmp_mode_collapse}
        \end{figure}

    \subsection{Main Results}

        \textbf{Can marginal models be fine-tuned into consistent multi-agent predictors?}
        To answer this question, we apply SimPO on marginal models in multi-agent scenarios. Table \ref{tab:main_results_marginal} shows our results: pSCR improves significantly (up to 44\% for FJMP-Marginal and 57\% for QCNet in AV2), indicating that modes with collisions are far less probable after SimPO.
        The decrease of SCR (7\% for FJMP-Marginal and 27\% for QCNet in AV2, 9\% for FJMP-Marginal in Interaction) is evidence that the probability updates due to the SimPO objective in (\ref{eq:simpo_tp_loss}) change both agent pairings and trajectory shapes in such a way that the total number of collisions decreases, and not merely their predicted likelihood of occurring. We also notice that, in all cases, distance-metrics like MinJointFDE$_6$ are degraded, resulting in a trade-off between consistency and accuracy; the degradation however is small compared to the improvement in collision rates.

        SimPO effectively mitigates collisions in interactive scenarios, as shown in Fig. \ref{fig:k6_qcnet_qualitative_case}. By updating probabilities, it adjusts agent pairings and trajectory shapes to prioritize collision-free modes, such as having agent 1 yield to agent 2, thereby reducing preference costs and improving consistency. FJMP-Marginal shows smaller improvements across the entire validation dataset of AV2 compared to QCNet; however, on \textit{interactive} validation scenes of AV2 (selected as explained in Section \ref{sec:pref_data_extraction}), it shows an improvement  of 29\% in SCR and 56\% in pSCR.

        \textbf{Can joint models have their consistency further improved by SimPO?}
        Table \ref{tab:main_results_joint} presents the results of applying SimPO on joint models. Notice that BeTop, a recent state-of-the-art model, shows significant improvements in both SCR and pSCR (36\% and 37\%, respectively) in the interactive validation split of WOMD, while suffering from only a 7\% degradation in MinJointFDE$_6$. Much like the behavior observed with QCNet on AV2, for example, agent-level predictions are aggregated into joint modes with fewer and less likely collisions.
        
        The less pronounced improvements of FJMP-Joint on AV2 suggest once again a trade-off: FJMP-Marginal achieves worse accuracy in terms of MinJointFDE$_6$ than the full Joint model, but has a simpler training pipeline and seems to benefit more from SimPO, achieving lower values of pSCR than its Joint counterpart after fine-tuning. What to prioritize between accuracy, consistency and computational expenses ultimately depends on downstream requirements in the autonomous vehicle pipeline.
        
        \textbf{Can oversampling trajectories before applying SimPO further improve collision rates?}
        As mentioned, the large decrease in pSCR observed in Tables \ref{tab:main_results_marginal} and \ref{tab:main_results_joint} suggests that rearranging agent pairings to achieve consistent predictions is the main effect of applying SimPO. However, improvements in plain SCR are limited due to the small number of predictions available. This is visible in Fig. \ref{fig:k6_qcnet_qualitative_case}: although collisions are avoided by rearrangement, notice that agents still come unrealistically close to each other in mode 6, which is reflected in the preference cost. We experiment with making the model output more joint modes so that the increased diversity due to the new range of available behaviors will lead to agent pairings that have collisions being pushed out of the top-6 in the preference ranking.

        As shown in Fig. \ref{fig:oversampling_scr}, this indeed occurs: notice that as $K$ increases, SCR improvement also gets larger. For instance, using QCNet on AV2 and predicting 15 joint modes, we obtain a decrease of 64\% in SCR. This means that the greater availability of behaviors indeed causes inconsistent modes to be pushed out of the top six with the probability updates and the subsequent changes in agent pairings. Fig. \ref{fig:k15_qcnet_qualitative_case} shows that after SimPO, unrealistic joint behaviors such as mode 6 in the left scenario or mode 5 in the right scenario are fixed and the collision is avoided, without significantly compromising trajectory shapes.

        \textbf{How does mode-collapse affect the results of SimPO fine-tuning?}
        Finally, our oversampling results assume that new behaviors will be available as $K$ increases, allowing the formation of new consistent joint modes. What happens when this is not the case? Notice that in both Tables \ref{tab:main_results_marginal} and \ref{tab:main_results_joint}, FJMP-Marginal and FJMP-Joint also present improvements in collision rates, but smaller ones. Note also that the trade-off in favor of consistency is less pronounced in the Interaction dataset for FJMP-Joint. This is because in our experiments both models suffer heavily from mode collapse, as shown in Fig. \ref{fig:fjmp_mode_collapse}. This is a known problem of decoding multiple futures directly from encoded agent features \cite{shi2024mtr++}.  Mode collapse makes it so that the small changes in trajectory shapes or rearrangement of agent pairings coming from SimPO are sometimes not enough to avoid collisions in many cases.
        
        This finding underscores the importance of developing models capable of covering all possible maneuvers, not only to enhance planning by providing autonomous vehicles access to diverse scenarios for safer route proposals but also to enable the application of techniques like SimPO for further reductions in predicted collisions. Analogous to LLMs, where access to the full distribution of text generations allows for efficient guidance via preference optimization, a comprehensive distribution of trajectory predictions facilitates similar improvements in scene consistency. To this effect anchor or intention point-based decoding strategies such as is used in QCNet and BeTop seem more promising.

    \subsection{Supplementary Results and Ablation Studies}
    \label{sec:ablation}


        \textbf{Rewards of best modes indeed increase.}
        We observe in our experiments a consistent increase in the rewards of best modes compared to rewards of worst modes in the preference ranking during the fine-tuning step. This means that the top mode, which due to the preference ranking will most often not have collisions, has its predicted likelihood increased compared to that of the worst mode, which will more often have collisions. SimPO can therefore effectively decrease the predicted likelihood of modes with collisions; this is further supported by the results in Tables \ref{tab:main_results_marginal} and \ref{tab:main_results_joint}.

        \textbf{Optimizing the preference cost directly leads to unrealistic shapes.}
        The availability of an automatically measurable and differentiable preference cost prompts consideration of whether using it directly could yield comparable improvements in collision rates, be it by (i) directly incorporating the preference cost into the loss function during standard trajectory prediction pretraining, or (ii) utilizing the preference cost as the sole objective in a subsequent fine-tuning stage. Regarding approach (i), pretraining loss functions are often tailored to specific model architectures, suggesting that adding this cost term might lack the consistency and broad applicability of a dedicated fine-tuning step applicable to various models. Concerning approach (ii), our experiments revealed that directly optimizing the preference cost leads to degenerate solutions, specifically unrealistic predictions where trajectories erroneously collapse onto the agents.

        \textbf{The target reward margin adaptation is crucial.} 
        Table \ref{tab:gamma_ablation} shows that, much like what is found in the original SimPO paper \cite{meng2024simpo}, proper tuning of the target reward margin is essential to achieve the desired improvements. Not using it at all, i.e. setting $\gamma = 0$, leads to catastrophic deterioration of both collision rates and accuracy. By forcing the reward difference among the ranked modes to be higher, we are able to achieve improvements in both SCR and pSCR, and our experiments suggest that there is an ideal interval of $\gamma$ in which accuracy is minimally degraded. This also demonstrates that the particular adaptation to the target reward margin implementation shown in (\ref{eq:simpo_tp_loss}) where the reward difference scales with the relative distance in ranking was helpful to achieve the desired result.

        \begin{table}[t]
            \caption{Effect of $\gamma$ on collision rate. Experiments on QCNet, $K=6$, $\beta = 2$. Collision rates are multiplied by $10^{3}$.}
            \label{tab:gamma_ablation}
            \centering
            \begin{tabular}{@{}c|lll@{}}
            \hline
            $\gamma$ & SCR     & pSCR  & MinJointFDE$_6$   \\
            \hline \hline
            Before SimPO & 8.64 & 2.86 & 1.366 \\
            \hline
            0.0        & 28.75 \textcolor{red}{\scriptsize (+233\%)} & 16.52 \textcolor{red}{\scriptsize (+478\%)} & 1.741 \textcolor{red}{\scriptsize (+27\%)} \\
            2.0        & 9.87 \textcolor{red}{\scriptsize (+14\%)} & 2.07 \textcolor{blue}{\scriptsize (-28\%)} & 1.373 \textcolor{red}{\scriptsize (+1\%)} \\
            5.0        & 6.29 \textcolor{blue}{\scriptsize (-27\%)} & 1.23 \textcolor{blue}{\scriptsize (-57\%)} & \textbf{1.375} \textcolor{red}{\scriptsize (+1\%)} \\
            7.0        & 5.92 \textcolor{blue}{\scriptsize (-31\%)} & \textbf{0.81} \textcolor{blue}{\scriptsize (-72\%)} & 1.401 \textcolor{red}{\scriptsize (+3\%)} \\
            10.0        & \textbf{5.41} \textcolor{blue}{\scriptsize (-37\%)} & 0.87 \textcolor{blue}{\scriptsize (-70\%)} & 1.450 \textcolor{red}{\scriptsize (+6\%)} \\
            \hline
            \end{tabular}
        \end{table}

        \begin{table}[t]
            \caption{Effect of $\lambda$ on collision rate. Experiments on QCNet, $K=6$, $\beta=2$, $\gamma = 5$. Collision rates are multiplied by~$10^{3}$.}
            \centering
            \begin{tabular}{@{}c|lll@{}}
            \hline
            $\lambda$ & SCR     & pSCR & MinJointFDE$_6$   \\ \hline \hline
            Before SimPO & 8.64 & 2.86 & 1.366 \\
            \hline
            0        & \textbf{6.02} & 1.30  & 1.381  \\
            $10^3$        & 6.29 & \textbf{1.23} & \textbf{1.366}  \\ \hline
            \end{tabular}
            
            \label{tab:lambda_ablation}
        \end{table}

        \textbf{The design of the preference metric is flexible.} 
        As shown in Table \ref{tab:lambda_ablation}, using $\lambda = 0$ can also lead to further improvements in SCR and pSCR, at the cost of worse final values of distance-based metrics. This suggests that simply using the avgFDE as a preference metric can be beneficial, depending on if the emphasis of the end user is on consistency or accuracy. Although avgFDE presupposes knowledge of future trajectories, we stress that any metric can in principle be used in order to rank modalities and apply SimPO, including metrics for which such knowledge is not required. We choose to present the main results in Tables \ref{tab:main_results_marginal} and \ref{tab:main_results_joint} using $\lambda = 10^3$ since we try to minimize any loss in accuracy while guarding large improvements in collision rates.

\section{Conclusion}
\label{sec:concl}

    This work is, to our knowledge, the first to apply preference optimization on trajectory prediction models, inspired by its success in guiding the output of LLMs. Both LLMs and trajectory prediction models share features like the inherent multi-modal uncertainty in their outputs—whether texts or trajectories—and the potential to rank these outputs based on preferences. Leveraging these similarities, we apply preference optimization to improve the consistency of joint predictions, by adapting the SimPO loss for trajectory prediction and using preference rankings generated automatically. Our results show that preference optimization improves the consistency of multi-agent trajectory predictions with minimal impact on accuracy, suggesting a promising new avenue for enhancing predictions to better meet the requirements of the AV pipeline, and which can be in principle applied to any model.

    \paragraph*{Limitations and future work} Our particular implementation of SimPO for trajectory prediction models used a preference cost designed to improve specifically scene consistency, whereas other target points of improvement could be studied. Another direction for further exploration would be on how the use of different rewards or preference metrics to rank different modes affect final collision rates and distance-based metrics.

\bibliographystyle{IEEEtran}
\bibliography{biblio}

\end{document}